\documentclass{article}

\usepackage{microtype}
\usepackage{graphicx}
\usepackage{subfigure}
\usepackage{booktabs} 

\usepackage{hyperref}

\usepackage[accepted]{icml2021}

\icmltitlerunning{Emulating Aerosol Microphysics with a Machine Learning}

\begin{document}

\twocolumn[
\icmltitle{Emulating Aerosol Microphysics with Machine Learning}



\icmlsetsymbol{equal}{*}

\begin{icmlauthorlist}
\icmlauthor{Paula Harder}{itwm,tuk,mlc}
\icmlauthor{Duncan Watson-Parris}{ox}
\icmlauthor{Dominik Strassel}{itwm}
\icmlauthor{Nicolas Gauger}{tuk}
\icmlauthor{Philip Stier}{ox}
\icmlauthor{Janis Keuper}{itwm,imla}
\end{icmlauthorlist}

\icmlaffiliation{itwm}{Competence Center High Performance Computing, Fraunhofer Institute for Industrial Mathematics, Kaiserslautern, Germany}
\icmlaffiliation{ox}{Dpt. of Atmospheric, Oceanic and Planetary Physics, University of Oxford, UK}
\icmlaffiliation{imla}{Institute for Machine Learning and Analytics (IMLA), Offenburg University, Germany}
\icmlaffiliation{mlc}{Fraunhofer Center Machine Learning, Germany}
\icmlaffiliation{tuk}{Scientic Computing, University of Kaiserslautern, Kaiserlautern, Germany}

\icmlcorrespondingauthor{paula.harder@itwm.fraunhofer.de}

\vskip 0.3in
]
\printAffiliationsAndNotice{}

\begin{abstract}
Aerosol particles play an important role in the climate system by absorbing and scattering radiation and influencing cloud properties. They are also one of the biggest sources of uncertainty for climate modeling. Many climate models do not include aerosols in sufficient detail. In order to achieve higher accuracy, aerosol microphysical properties and processes have to be accounted for. This is done in the ECHAM-HAM global climate aerosol model using the M7 microphysics model, but increased computational costs make it very expensive to run at higher resolutions or for a longer time. We aim to use machine learning to approximate the microphysics model at sufficient accuracy and reduce the computational cost by being fast at inference time. The original M7 model is used to generate data of input-output pairs to train a neural network on it. By using a special logarithmic transform we are able to learn the variables tendencies achieving an average $R^2$ score of $89\%$. On a GPU we achieve a speed-up of 120 compared to the original model.
\end{abstract}

\section{Introduction}
Aerosol forcing remains the largest source of uncertainty in the anthropogenic effect on the current climate \cite{https://doi.org/10.1029/2019RG000660}. The aerosol cooling effect hides some of the positive radiative forcing caused by greenhouse gas emissions and future restrictions to lower air pollution might result in stronger observed warming. Aerosols impact climate change through aerosol-radiation interactions and aerosol-cloud interactions \cite{IPCCWG1PhysicalStocker2013}. They can either scatter or absorb radiation, which depends on the particle's compounds. Black carbon aerosols from fossil fuel burning for example have a warming effect by absorbing radiation and whereas sulphate from volcanic eruptions has a cooling effect by scattering radiation. Clouds strongly influence the earth's radiation budget by reflecting sunlight and aerosols can change cloud properties significantly by acting as cloud condensation nuclei (CCN). A higher concentration of aerosols leads to more CCN which, for a fixed amount of water, results in more but smaller cloud droplets. Smaller droplets increase the cloud's albedo \cite{TWOMEY19741251} and can enhance the cloud's lifetime \cite{ALBRECHT1227}. 

Many climate models often consider aerosols only as external parameters, they are read once by the model, but then kept constant throughout the whole model run. In the case where some aerosol properties are modeled, there might be no distinction between aerosol types, and just an overall mass is considered. To incorporate a more accurate description of aerosols, aerosol-climate modeling systems like ECHAM-HAM \cite{gmd-12-1643-2019} have been introduced. It couples the ECHAM General Circulation Model (GCM) with a complex aerosol model called HAM. The microphysical core of HAM is either SALSA \cite{acp-8-2469-2008} or the M7 model \cite{m7}. We consider the latter here. M7 uses seven log-normal modes to describe aerosol properties, the particle sizes are represented by four size modes, nucleation, aitken, accumulation, and coarse, of which the aitken, accumulation, and coarse can be either soluble or insoluble. It includes processes like nucleation, coagulation, condensation, and water uptake, which lead to the redistribution of particle numbers and mass among the different modes. In addition, M7 considers five different components - Sea salt (SS), sulfate (SO4), black carbon (BC), primary organic carbon (OC), and dust (DU). M7 is applied to each grid box independently, it does not model any spatial relations.

More detailed models come with the cost of increased computational time: ECHAM-HAM can be run at 150km resolution for multiple decades. But to run for example storm-resolving models the goal is ideally a 1km horizontal grid resolution and still be able to produce forecasts up to a few decades. If we want to keep detailed aerosols descriptions for this a massive speedup of the aerosol model is needed. 

Replacing climate model components with machine learning approaches and therefore decreasing a model's computing time has been shown to be successful in the past. There are several works on emulating convection, both random forest approaches \cite{gorman} and deep learning models \cite{Rasp9684}, \cite{gentine}, \cite{beucler2020physicallyconsistent} have been explored. Recently, multiple consecutive neural networks have been used to emulate a bin microphysical model for warm rain processes \cite{gettelmann}. Silva et al. \cite{gmd-2020-393} compare several methods, including deep neural networks, XGBoost, and ridge regression as physically regularized emulators for aerosol activation. In addition to the aforementioned approaches random forest approaches have been used to successfully derive the CCN from atmospheric measurements \cite{acp-20-12853-2020}.

Our work shows a machine learning approach to approximate the M7 microphysics module. We investigated different approaches, neural networks as well as ensemble models like random forest and gradient boosting, with the aim of achieving the desired accuracy and computational efficiency, finding the neural network appeared to be the most successful. We use data generated from a realistic ECHAM-HAM model run and train a model offline to predict one time step of the aerosol microphysics. The underlying data distribution is challenging, as the changes in the variables are often zero or very close to zero. We do not predict the full values, but the tendencies, which requires a logarithmic transformation for both positive and negative values. After we considered different hyperparameters, a 2-layer, fully connected network to solve this multivariate regression problem is chosen. To train, validate and test our model in a representative way, we use about 11M data points from separate days spread out through the year. On the test set, an overall regression coefficient of $89\%$ and a root mean squared error (RMSE) of $24.9\%$ are achieved. For $75\%$ of the variables, our model's prediction has a $R^2$ value of over $95\%$ on the logarithmically transformed tendencies. Using a GPU we achieve a huge speed-up of 120 compared to the original model, on a single CPU we are still 1.6 faster.

\section{Methodology}

\subsection{Data}
\subsubsection{Data generation}
To easily generate data we extract the aerosol microphysics model from the global climate model ECHAM-HAM and develop a stand-alone version. To obtain input data ECHAM-HAM is run for four days within a year. We use a horizontal resolution of 150km at the equator, overall 31 vertical levels, 96 latitudes, and 192 longitudes. A time step length of 450s is chosen. This yields a data set of over 100M points for each day and we only use a subset of 5 times a day, resulting in about 2.85M points per day. We use the days in January and April for training, a day in July for validation, and the October data for testing the model. Because the two sea salt variables only change in $0.2\%$ of the time we do not model them here. The masses and particle concentrations for different aerosol types and different modes are both inputs and outputs, the output is the value one time step later. Atmospheric state variables like pressure, temperature, and relative humidity are used as inputs only. A full list of input and output values can be found in the appendix in Table \ref{variables}.

\subsubsection{Data distribution and transformation}
Compared to the full values the tendencies are very small, so we aim to predict the tendencies, not the full values, where it is possible. Depending on the variable we have very different size scales, but also a tendency for a specific variable may span several orders of magnitude. In some modes, variables can only grow, in some only decrease, and in others do both. Often the majority of tendencies are either zero or very close to zero, but a few values might be very high. Usual normalizing would not solve the problem, that the training and evaluation would be dominated by a small number of huge values. As we want to give more weight to the prominent case, the smaller changes, a log transform is necessary, as for example done in \cite{gettelmann}. To have a continuous transformation, being able to deal with negative values, and only using one network and transformation we choose the following

$y = \left\{
\begin{array}{ll}
\log{\sqrt{x}+1} & x \geq 0 \\
-\log{\sqrt{-x}+1} & \, \textrm{else.} \\
\end{array}
\right. $

The additional square root spreads out the values close to zero. The above transformation is applied to all variables. After the logarithmic transformation, we normalize and standardize all input and output values.
\subsection{Model}

For emulating the M7 microphysics model we explored different machine learning approaches, including random forest regression, gradient boosting, and a simple neural network. The neural network approach appears to be the most successful (See Figure \ref{compare}) for this application and will be presented here.

\subsubsection{Network architecture}
We employ a fully connected network with a sigmoid activation function and two hidden layers, each hidden layer contains 256 nodes. Using zero hidden layers results in a linear regression and does not have the required expressivity for our task, one hidden layer already achieve a good performance, after two layers the model we could not see any further improvements. The width of layers improves performance up to about 256 nodes and then stagnates as well. Even though the network's performance is not very sensitive to the choice of activation function, we achieve the best results using a sigmoid function.

\subsubsection{Training}
We train the network using the Adam optimizer with a learning rate of $10^{-3}$, a weight decay of $10^{-9}$, and a batch size of 4096. Sensitivity analysis to different learning rates, weight decays, and batch size values can be found in the appendix. As a loss function, we choose an MSE loss and use early stopping with a patience of 10 epochs, where our training process stops after about 60 epochs. The training on a single NVIDIA Titan V GPU takes about 2 hours.

\section{Results}

\subsection{Predictive performance}

\begin{figure}[ht]
\vskip 0.2in
\begin{center}
\centerline{\includegraphics[width=\columnwidth]{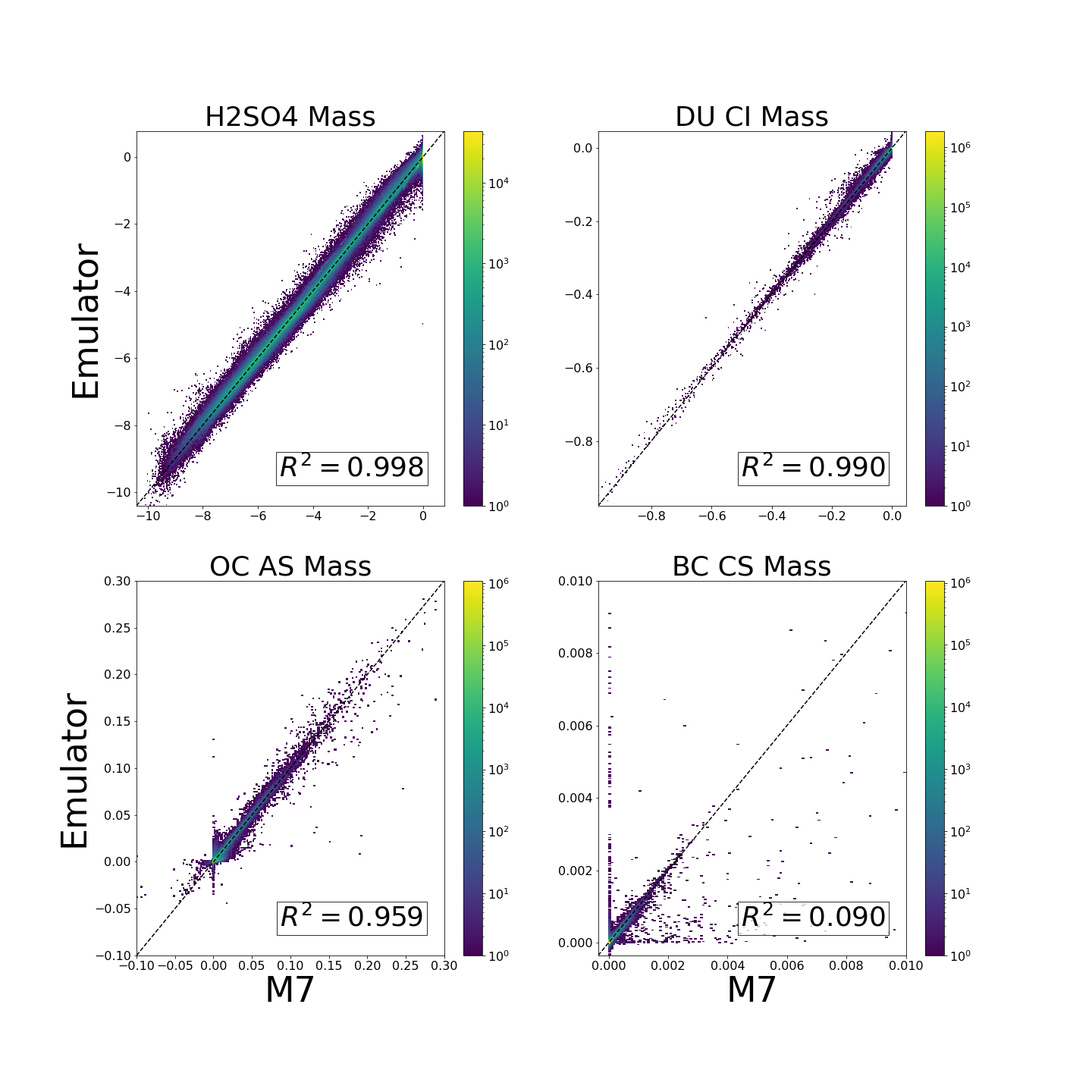}}
\caption{Predicted value by emulator versus true value from M7 model, tendencies, logarithmically transformed.}
\label{four}
\end{center}
\vskip -0.2in
\end{figure} 

To evaluate the performance of our emulator we run the neural network on our independent test set. We calculate the regression coefficient, the $R^2$ score, on the log-transformed tendencies. In order to be more comparable, the different scales for the different variables suggest a normalized RMSE, therefore we calculate this metric on the normalized values. Table \ref{overal_performance} shows the average performance over all 28 predicted variables, the masses, and concentration after one time step. The performance does not drop significantly compared to the training set, but there is a decrease in the $R^2$ value from the validation set to the test case. Test scores for all variables separately can be found in Table \ref{scores_all}. For 21 out of 28 variables $R^2$ scores of over $95\%$ are achieved. We found that two variables seem problematic to model, which is organic carbon in the aitken mode and especially black carbon in the coarse mode. The bad performance for black carbon was not observed on the validation set, which could indicate some seasonal differences. A possible solution could be to train on a full year and evaluate on the next.

\begin{table}[t]
\caption{Regression performance of M7 emulator.}
\label{sample-table}
\vskip 0.15in
\begin{center}
\begin{small}
\begin{sc}
\begin{tabular}{lcr}
\toprule
Case & RMSE & $R^2$  \\
\midrule
train     & 0.230& 0.933\\
val &0.254&0.924\\
test & 0.249&0.892 \\
\bottomrule
\end{tabular}
\end{sc}
\end{small}
\end{center}
\vskip -0.1in
\label{overal_performance}
\end{table}

\begin{table}[t]
\caption{Regression performance of the M7 emulator for all variables.}
\label{sample-table}
\vskip 0.15in
\begin{center}
\begin{small}
\begin{sc}
\begin{tabular}{lccr}
\toprule
Variable & $R^2$ & RMSE  \\
\midrule
H2SO4 mass    & 0.9987& 0.043 \\
SO4 ns mass & 0.967& 0.187 \\
SO4 ks mass & 0.837& 0.369 \\
SO4 as mass & 0.961& 0.186 \\
SO4 cs mass & 0.985& 0.116 \\
bc ks mass &0.765& 0.457 \\
bc as mass & 0.969& 0.174 \\
bc cs mass & 0.090& 0.523 \\
bc ki mass & 0.990& 0.010 \\
oc ks mass & 0.477& 0.652 \\
oc as mass & 0.959& 0.213 \\
oc cs mass & 0.631& 0.220 \\
oc ki mass & 0.989& 0.117 \\
du as mass & 0.980& 0.065 \\
du cs mass & 0.989& 0.063 \\
du ai mass & 0.988& 0.050 \\
du ci mass & 0.990& 0.062 \\
ns concentration& 0.977& 0.154 \\
ks concentration & 0.823& 0.329 \\
as concentration & 0.699& 0.512 \\
cs concentration & 0.985& 0.074 \\
ki concentration & 0.993& 0.088 \\
ai concentration & 0.984& 0.058 \\
ci concentration & 0.987& 0.067 \\
ns water& 0.995& 0.068 \\
ks water & 0.997& 0.041\\
as water& 0.991& 0.096\\
cs water & 0.993& 0.082\\
\bottomrule
\end{tabular}
\end{sc}
\end{small}
\end{center}
\vskip -0.1in
\label{scores_all}
\end{table}

In Figure \ref{four} we show the emulator tendency prediction plotted against the M7 module predictions for four variables. We show the best and worst-performing, as well as the variables closest to 25 and 75 percentiles, plots for all variables, can be found in the appendix, in Figure \ref{all}. In the plots we can observe that often there are stronger deviations at zero, the model struggles with detecting whether a variable is changing at all or not. For black carbon coarse mode, we can see, that even though we applied a log transform, most of the values are very close to zero, whereas a few a many orders of magnitude bigger. This probably makes it hard for the model to learn the dependencies. The scores for back-transformed values show weaker performance for some variables, looking at the scores for the full variables though a perfect $R^2$ is achieved apart from three variables, see Table \ref{back-table}.

Although the model performs well in an offline evaluation, it still remains to be shown how it performs when plugged back into the GCM and run for multiple time steps, good offline performance is no guarantee for good online performance \cite{gmd-13-2185-2020} in the case of convection parameterization, where a model crash could occur. In our case, it is likely that offline performance is a good indicator for online performance, as a model crash is not expected, because aerosols do not strongly affect large-scale dynamics.

\subsection{Runtime}

We conduct a preliminary runtime analysis by comparing the Python runtime for the emulator with the Fortran runtime of the original model. For the M7 model, the 31 vertical levels are calculated simultaneously and for the emulator, we predict one time step at once globally. We use a single NVIDIA Titan V GPU and single Intel Xeon 4108 CPU. As shown in Table \ref{runtime} we can achieve a massive speed-up of 120 using a GPU compared to the original model, but using a single CPU we are only 1.6 times faster. Further speed-ups could be achieved by using multiple CPUs, a smaller network architecture, and an efficient implementation in Fortran.

\begin{table}[t]
\caption{Runtime comparison for the original M7 model and the M7 emulator.}
\label{runtime}
\vskip 0.15in
\begin{center}
\begin{small}
\begin{sc}
\begin{tabular}{lccr}
\toprule
Model & M7 & Emulator GPU & Emulator CPU  \\
\midrule
time (s)    & 5.781& 0.048 &3.716\\
speed-up & - & 120.4 & 1.6\\
\bottomrule
\end{tabular}
\end{sc}
\end{small}
\end{center}
\vskip -0.1in
\label{overal_performance}
\end{table}

\section{Conclusion and future work}
This work shows how neural networks can be used to learn the mapping of an aerosol microphysics model. Our model approximates the log-transformed tendencies well for nearly all variables and excellent for most of them. Using a GPU we achieve a significant speed-up.

How much of a speed-up can be achieved in the end remains to be shown, when the model is used in a GCM run. For long-time climate modeling, it is important that our model does not show any mass biases and slowly loses or gains mass. To prevent that it has to be investigated if mass conservation is systematically violated or if it is symmetrically distributed around zero. To enforce mass conservation one possibility is to add a penalty term to the network's loss function. Although it is not guaranteed, it could decrease the violation of mass conservation. In some cases, our model might also predict nonphysical values like negative masses and a mass fixer could be necessary. To employ the full potential of deep learning it could be possible to include spatial relationships in the model and not consider each grid box independently.

\section*{Acknowledgements}
DWP and PS acknowledge funding from NERC project NE/S005390/1 (ACRUISE) as well as from the European Union’s Horizon 2020 research and innovation programme iMIRACLI under Marie Skłodowska-Curie grant agreement No 860100. PS additionally acknowledges support from the ERC project RECAP and the FORCeS project under the European Union’s Horizon 2020 research programme with grant agreements 724602 and 821205. We would like to thank the reviewers for their helpful comments.
\bibliography{aerosol}
\bibliographystyle{icml2021}
\newpage
\appendix
\section{Appendix}

\begin{table}[h]
\caption{Regression performance of different machine learning approaches. Average scores over all predicted variables}
\label{compare}
\vskip 0.15in
\begin{center}
\begin{small}
\begin{sc}
\begin{tabular}{lcr}
\toprule
Method & RMSE & $R^2$  \\
\midrule
Random Forest     & 0.9570& -2.139\\
Gradient Boosting &1.36&-11.197\\
Neural Network & 0.249&0.892 \\
\bottomrule
\end{tabular}
\end{sc}
\end{small}
\end{center}
\label{overal_performance}
\end{table}
\vspace{-0.2in}
\begin{table}[!h]
\caption{Modelled variables.}
\label{sample-table}
\begin{center}
\begin{small}
\begin{sc}
\begin{tabular}{lccr}
\toprule
Variable & Unit & Input & Output  \\
\midrule
Pressure & $\mbox{Pa}$ &$\surd$ & \\
Temperature & K&$\surd$& \\
Rel. Humidity & - &$\surd$ & \\
ionization rate & - &$\surd$  & \\
cloud cover & - &$\surd$ & \\
Boundary layer & - &$\surd$ & \\
Forest fraction & - &$\surd$ & \\
H2SO4 prod. rate  & $cm^{-3}s^{-1}$ & $\surd$&  \\
SS as mass    & $\mu g\  m^{-3}$&$\surd$&   \\
ss cs mass   & $\mu g\  m^{-3}$&$\surd$&   \\
H2SO4 mass    & $\mu g\  m^{-3}$ &$\surd$& $\surd$ \\
SO4 ns mass & $molec.\  m^{-3}$&$\surd$& $\surd$ \\
SO4 ks mass & $molec.\  m^{-3}$&$\surd$& $\surd$ \\
SO4 as mass & $molec.\  m^{-3}$&$\surd$& $\surd$ \\
SO4 cs mass & $molec.\  m^{-3}$&$\surd$& $\surd$ \\
bc ks mass & $\mu g\  m^{-3}$&$\surd$& $\surd$ \\
bc as mass & $\mu g\  m^{-3}$&$\surd$& $\surd$ \\
bc cs mass & $\mu g\  m^{-3}$&$\surd$& $\surd$ \\
bc ki mass & $\mu g\  m^{-3}$&$\surd$&$\surd$ \\
oc ks mass & $\mu g\  m^{-3}$&$\surd$& $\surd$ \\
oc as mass & $\mu g\  m^{-3}$&$\surd$& $\surd$ \\
oc cs mass & $\mu g\  m^{-3}$&$\surd$& $\surd$ \\
oc ki mass & $\mu g\  m^{-3}$&$\surd$& $\surd$ \\
du as mass &$\mu g\  m^{-3}$&$\surd$& $\surd$ \\
du cs mass & $\mu g\  m^{-3}$&$\surd$& $\surd$ \\
du ai mass & $\mu g\  m^{-3}$&$\surd$& $\surd$ \\
du ci mass & $\mu g\  m^{-3}$& $\surd$& $\surd$ \\
ns concentration&$ cm^{-3}$ &$\surd$& $\surd$ \\
ks concentration & $ cm^{-3}$ &$\surd$& $\surd$ \\
as concentration & $ cm^{-3}$ &$\surd$& $\surd$ \\
cs concentration & $ cm^{-3}$ &$\surd$& $\surd$ \\
ki concentration & $ cm^{-3}$ &$\surd$& $\surd$ \\
ai concentration & $ cm^{-3}$ &$\surd$& $\surd$ \\
ci concentration & $ cm^{-3}$ &$\surd$& $\surd$\\
ns water& $kg\ m^{-3} $& & $\surd$ \\
ks water & $kg\ m^{-3} $& & $\surd$ \\
as water& $kg\ m^{-3} $ & &$\surd$ \\
cs water& $kg\ m^{-3} $ & & $\surd$ \\
\bottomrule
\end{tabular}
\end{sc}
\end{small}
\end{center}
\vskip -0.1in
\label{variables}
\end{table}

\begin{table}[h]
\caption{Regression performance back-transformed values of the M7 emulator for all variables. Second column is the tendency in original units and third row the full values.}
\label{back-table}
\begin{center}
\begin{small}
\begin{sc}
\begin{tabular}{lccr}
\toprule
Variable & $R^2$ tend &  $R^2$ full  \\
\midrule
H2SO4 mass    & 0.683& 0.708 \\
SO4 ns mass & 0.000& 0.000 \\
SO4 ks mass & -3.9$\cdot 10^{42}$& -7.5$\cdot 10^{37}$\\
SO4 as mass & 0.259& 1.0\\
SO4 cs mass & -0.672& 1.0 \\
bc ks mass &0.938& 1.0 \\
bc as mass & 0.921& 1.0 \\
bc cs mass & -42.842&1.0 \\
bc ki mass & 0.937& 1.0 \\
oc ks mass & 0.624& 1.0 \\
oc as mass & 0.744& 1.0 \\
oc cs mass & 0.691& 1.0 \\
oc ki mass & 0.731& 1.0 \\
du as mass & 0.898& 1.0 \\
du cs mass & 0.991& 1.0 \\
du ai mass & 0.992& 1.0\\
du ci mass & 0.992& 1.0\\
ns concentration& 0.031& 0.031 \\
ks concentration & 0.250&1.0 \\
as concentration & 0.869& 1.0 \\
cs concentration & 0.985& 1.0 \\
ki concentration &0.781& 1.0 \\
ai concentration & 0.984& 1.0\\
ci concentration & 0.989& 1.0 \\
ns water& 0.863& 0.863 \\
ks water & 0.994& 0.994\\
as water& 0.933& 0.933\\
cs water & 0.975& 0.975\\
\bottomrule
\end{tabular}
\end{sc}
\end{small}
\end{center}
\vskip -0.1in
\label{scores_all}
\end{table}

\begin{figure*}[ht]
\vskip 0.2in
\begin{center}
\centerline{\includegraphics[width=16cm]{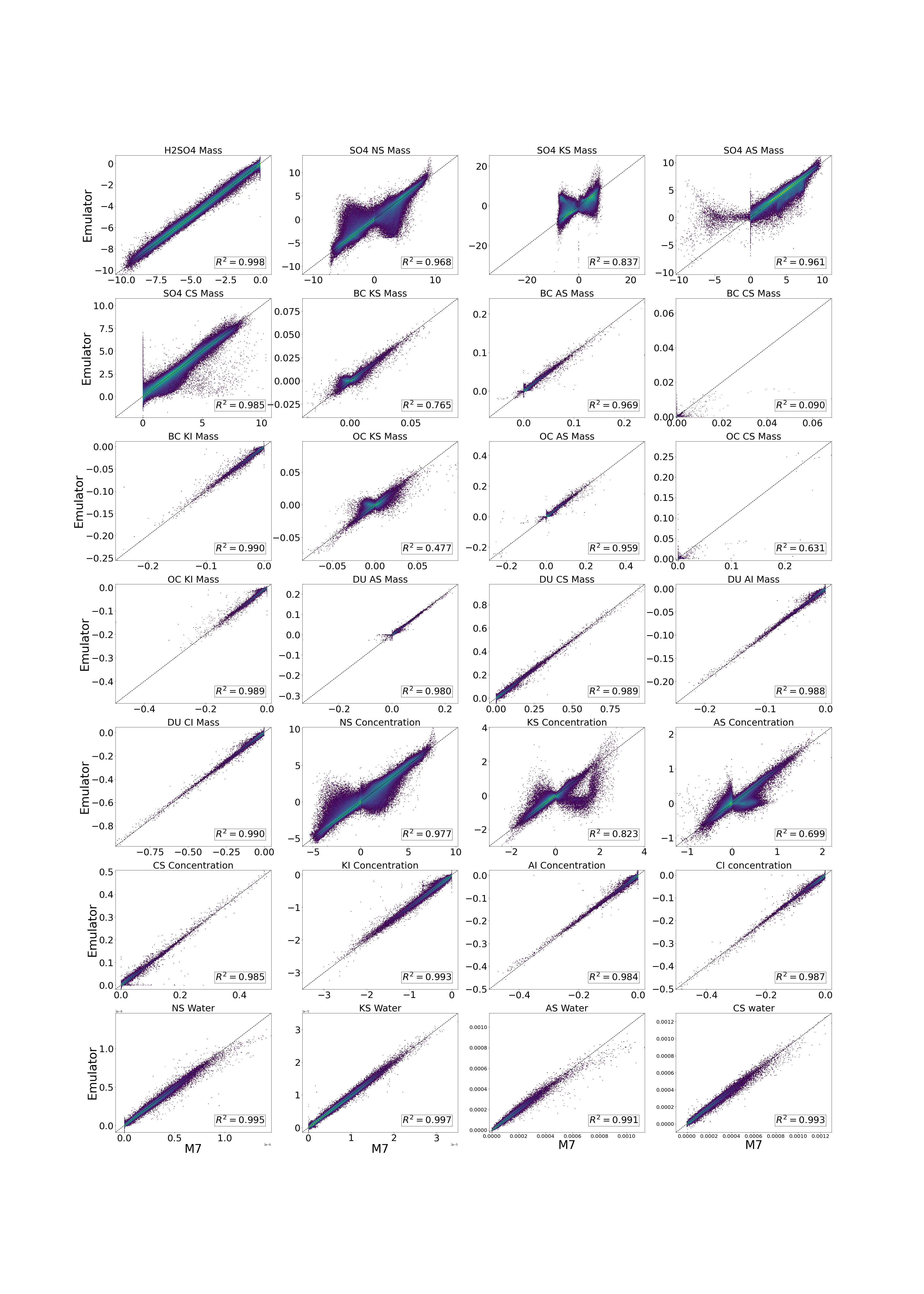}}
\caption{Predicted value by emulator versus true value from M7 model, tendencies, logarithmically transformed.}
\label{all}
\end{center}
\vskip -0.2in
\end{figure*}

\end{document}